\documentclass[10pt, a4paper]{article}
\usepackage{lrec2022} %
\usepackage{multibib}
\newcites{languageresource}{Language Resources}
\usepackage{graphicx}
\usepackage{tabularx}
\usepackage{soul}
\usepackage{titlesec}
\titleformat{\section}{\normalfont\large\bfseries\center}{\thesection.}{1em}{}
\titleformat{\subsection}{\normalfont\SmallTitleFont\bfseries\raggedright}{\thesubsection.}{1em}{}
\titleformat{\subsubsection}{\normalfont\normalsize\bfseries\raggedright}{\thesubsubsection.}{1em}{}
\renewcommand\thesection{\arabic{section}}
\renewcommand\thesubsection{\thesection.\arabic{subsection}}
\renewcommand\thesubsubsection{\thesubsection.\arabic{subsubsection}}

\usepackage{epstopdf}
\usepackage[utf8]{inputenc}

\usepackage{hyperref}
\usepackage{xstring}

\usepackage{color}

\usepackage[capitalise,noabbrev]{cleveref}
\usepackage{enumitem}
\usepackage{pifont}

\usepackage[normalem]{ulem} %

\newcommand{\toolname}{ALIGNMEET}
\newcommand{\xmark}{\ding{55}}%

\title{\toolname: A Comprehensive Tool for Meeting Annotation, Alignment, and Evaluation}
\name{Peter Pol\'ak, Muskaan Singh, Anna Nedoluzhko, Ond\v{r}ej Bojar} 

\address{
Charles University,
Faculty of Mathematics and Physics, Institute of Formal and Applied Linguistics \\
\texttt{\{polak,singh,nedoluzhko,bojar\}@ufal.mff.cuni.cz}}

\abstract{
Summarization is a challenging problem, and even more challenging is to manually create, correct, and evaluate the summaries. The severity of the problem grows when the inputs are multi-party dialogues in a meeting setup. To facilitate the research in this area, we present \toolname{}, a comprehensive tool for meeting annotation, alignment, and evaluation. The tool aims to provide an efficient and clear interface for fast annotation while mitigating the risk of introducing errors. Moreover, we add an evaluation mode that enables a comprehensive quality evaluation of meeting minutes. To the best of our knowledge, there is no such tool available. We release the tool as open source. It is also directly installable from  PyPI.
\\ \newline  }

\begin{document}

\maketitleabstract

\section{Introduction}

Meeting summarization into meeting minutes is primarily focused on topical coverage rather than on fluency or coherence. It is a challenging and tedious task, even when meeting summaries are created manually. The resulting summaries vary in the goals, style, and they are inevitably very subjective due to the human in the loop. Also, the awareness of the context of the meeting is essential to create adequate and informative summaries.
\subsection{Motivation}

First, there is a \textit{scarcity of large-scale meeting datasets}: There are a few meeting corpora, such as AMI \cite{mccowan2005ami} and ICSI \cite{janin2003icsi},  which are rather small,  on the order of a few dozens of hours each as represented in Table~\ref{tab:comparative}. Due to this fact, meeting summarization models are usually trained on news \cite{grusky2018newsroom}, stories \cite{hermann2015teaching}, Wikipedia \cite{frefel2020summarization,antognini2020gamewikisum}, and other textual corpora, relating poorly to meetings. 

\begin{table*}[ht]
\centering
\resizebox{0.99\textwidth}{!}{%
\begin{tabular}{llrrrrr}
\hline
\textbf{Category} &
\textbf{Dataset} &
  \textbf{\# Meetings} &
  \textbf{Avg Words (trans)} &
  \textbf{Avg Words (summ)} &
  \textbf{Avg Turns (trans)} &
  \textbf{Avg \# of speakers} \\\hline
Meeting & AutoMin (English) \cite{automin}  &     113   &       9,537   &   578     &     242   &  5.7  \\
 & AutoMin (Czech) \cite{automin}  &     53   &  11,784    &  292   &   579    &   3.6  \\
& ICSI \cite{janin2003icsi}             & 61      & 9,795 &   638   & 456    & 6.2 \\
& AMI \cite{mccowan2005ami}              & 137     & 6,970 & 179 & 335    & 4.0  \\\hline
Dialogue & MEDIASum \cite{zhu2021mediasum}   & 463,596 & 1,554  & 14   & 30   & 6.5 \\
& SAMSUM \cite{gliwa2019samsum}      & 16,369  & 84     & 20   & 10    & 2.2 \\
& CRD3 \cite{rameshkumar2020storytelling}        & 159     & 31,803 & 2,062 & 2,507 & 9.6 \\
& DiDi \cite{liu2019automatic}      & 328,880 & -        & -      & -      & 2.0   \\
& MultiWoz \cite{budzianowski2018multiwoz}   & 10,438  & 180    & 92   & 14   & 2.0  \\
\hline
\end{tabular}}%

\caption{Dialogue and meeting summarization datasets statistics. The number of words for dialogue, summary, turns, and speakers are averaged across the entire dataset. The meeting dataset statistics have been calculated and dialogue dataset statistics have been derived from \protect\newcite{zhu2021mediasum}.%
}
\label{tab:comparative}

\end{table*}
Second, when one tries to create such a collection or when a new meeting is to be processed, a \textit{reliable transcript} is needed, which is often impossible for the current automatic speech recognition systems (ASR). It usually requires a large amount of processing to make the transcript suitable for summarization.

Third, meeting transcripts are usually \textit{long text documents} consisting of multi-party dialogues (see \cref{tab:comparative}) with multiple topics. Moreover, meeting summaries are also longer compared to text summaries. The manifold structure and length of meeting transcripts and summaries make it difficult to traverse and follow the information for human annotators. Even training is difficult for a neural attention summarization model \cite{zhu2020hierarchical} with such input complexities. 

Finally, \textit{evaluation of meeting summarization} requires immediate access to the meeting transcript and sometimes even to the original sound recording to assess the quality of a particular summary point. The length of meeting transcripts and the amount of information quantity contained in a meeting itself easily cause a significant cognitive overload.

\subsection{Contribution}
We present an efficient, clean, and intuitive comprehensive alignment and evaluation tool which brings the following contributions:

\begin{itemize}
    \item An annotation platform for data creation and modification with multiple speaker support.
    \item Alignment between parts of a transcript with corresponding parts of summary.
    \item A novel evaluation strategy of meeting summaries which we integrate into the tool.
\end{itemize}

We release the tool as open source.\protect\footnote{\url{https://github.com/ELITR/alignmeet}} It is also directly installable from  PyPI.\footnote{\texttt{pip install alignmeet}} 

\section{Related Work}
This section studies existing \emph{annotation tools} and \emph{evaluation strategies} for meeting summarization.

\subsection{Annotation Tools}
\label{tools}
\cref{tab:annotation_tools} compares \toolname{} with other recent annotation tools for dialogue, conversation and meeting data. Most of the tools were designed for data curation. However, only some of them allow modifying the underlying datasets (see column~\emph{D}). Segmenting the dialogues or turns is possible in some tools (see column~\emph{A}) while speaker annotation is possible in almost all tools (column~\emph{B}). \toolname{} provides additional features of alignment and evaluation of meeting summaries.
 
\begin{table*}[ht]
\scriptsize
\centering
\begin{tabular}{llllllllll}
\hline
Tool & A & B & C & D & E & F & G & H\\\hline
\toolname{} (ours)  & \ding{51} & \ding{51} & \ding{51} & \ding{51} & \ding{51} & \ding{51} &   \ding{51} &   Python   \\
ELAN \cite{brugman2004annotating} &  \ding{51} & \ding{51} & \ding{51} & \ding{51} & \ding{51} &  &  \ding{51}   &     \\
EXMARaLDA \cite{schmidt2009exmaralda} &  \ding{51} & \ding{51} & \ding{51} & \ding{51} &  &  &  \ding{51}   &     \\
MATILDA \cite{cucurnia2021matilda} &  \ding{51} & \ding{51} & \ding{51} & \ding{51} & &   &  &    Python \\
metaCAT \cite{liu2020metacat}              &    \ding{51}                       & \ding{51}                              &  \ding{51} &            &                    &           &            &    Python \\
LIDA \cite{collins2019lida} & \ding{51} & \ding{51}      & \ding{51}               &                     &          &           & &  Python   \\
INCEpTion \cite{klie2018inception}  &   & \ding{51} &  & &    & & &   Java     \\
DOCCANO \cite{doccano}   &                          & \ding{51}                           &               &                    &           &   &  &          Python   \\
BRAT \cite{stenetorp2012brat}   &   & \ding{51} &   &   &   &   &  &         Python\\
NITE \cite{kilgour2006nite}                &                           &  \ding{51}                              &    \ding{51}           &    \ding{51}                &          &              & \ding{51} &  Java\\
SPAACy \cite{weisser2003spaacy}            &     \ding{51}                       &   \ding{51}                            &        \ding{51}       &                    &           &              & \ding{51}   &     Perl/Tk      \\
DialogueView \cite{heeman2002dialogueview} & \ding{51}                       & \ding{51}                          &               &                    &           &     &        &  Tcl/Tk  \\
ANVIL \cite{kipp2001anvil}   &  \ding{51} & \ding{51} & \ding{51}  & &  &    & \ding{51}  &   Java \\
NOMOS \cite{gruenstein2005meeting}   &  \ding{51}  &    \ding{51} & \ding{51}      & \ding{51}   &  &     & \ding{51}      &  Java\\
TWIST \cite{plussannotation2012} & \ding{51}  &  &  &  & & & &    -       \\\hline
\end{tabular}
\caption{Annotation Tool Comparison Table. Notation: A -- Turn/Dialogue Segmentation, B -- Edit Speaker Annotation, C -- Data Curation, D -- Data Modifications, E -- Alignment, F -- Evaluation,  G -- Audio/video playback, H -- Programming Language.}
\label{tab:annotation_tools}
\end{table*}

DialogueView \cite{heeman2002dialogueview} is a tool for annotation of dialogues with utterance boundaries, speech repairs, speech act tags, and discourse segments. It fails to capture inter-annotator reliability. 
TWIST \cite{plussannotation2012} is a tool for dialogue annotation consisting of turn segmentation and content feature annotation. The turn segmentation allows users to create new turn segments. Further, each segment can be labeled by selecting from a pre-defined feature list. 
This limits the user to pre-defined values. BRAT \cite{stenetorp2012brat} and DOCCANO \cite{doccano} are simple web-based annotation tools where you can only edit the dialogue and turns. 
BRAT also provides the user with automated recommendations. INCEpTion \cite{klie2018inception} is a platform for annotation of semantic resources such as entity linking. It provides automated recommendations to the user for annotation. NOMOS \cite{gruenstein2005meeting} is an annotation tool designed for corpus development and various other annotation tasks. Its main functionality includes multi-channel audio and video playback, compatibility with different corpora, 
platform independence and presentation of temporal, non-temporal, and related information. This tool is difficult to use by non-technical users and also lacks extensibility. ANVIL \cite{kipp2001anvil} allows multi-modal annotation of dialogues with the granularity in multiple layers of attribute-value pairs.  It also provides the feature of statistical processing but lacks the flexibility to add information to the annotation. NITE \cite{kilgour2006nite} is another multi-modal annotation tool aiding in corpora creation. The tool supports the time-alignment of annotation entities such as transcripts or dialogue structure. SPAACy \cite{weisser2003spaacy} is a semi-automated tool for annotating dialogue acts. It aids in corpus creation with tagging such as topic, mode, and polarity. In addition, it produces transcriptions in XML format that require a little post-editing. 
LIDA \cite{collins2019lida} is one of the most prominent tools for modern task-oriented dialogues with recommendations. However, LIDA does not support more than two speakers in the conversation or additional labeling (e.g., co-reference annotation). MATILDA \cite{cucurnia2021matilda} and metaCAT \cite{liu2020metacat} address some of the downsides. They add features such as inter-annotator disagreement resolution, customizable recommendations, multiple-language support, and user administration.  They still lack support for multiple speakers.

All these annotation tools provide annotation for dialogues, but for various textual phenomena. Our tool \toolname{} is specifically designed for meeting data creation or modification, alignment of meeting transcript regions with the corresponding summary items, and their evaluation. We also support dialogue and conversational datasets.

\subsection{Manual Evaluation}
\label{subsec:manual_evaluation}
Several researchers working on summarization have considered qualitative summary evaluation. The qualitative parameters include \emph{accuracy} \cite{Zechner01automaticsummarization,Zechner01automaticgeneration,8639531,10114532799813279987,Lai2013interspeech} which usually assesses the lexical similarity between produced text samples and the reference ones utilizing standard metrics such as BLEU \cite{papineni2002bleu} or ROUGE \cite{lin2004rouge}. The accuracy is easily computed in some of the applications when reference texts are available. \emph{Grammaticality} measures the capability of a model to produce grammatically correct texts \cite{liu2009extractive,mehdad2013abstractive}. It is mostly assessed by counting the different types of errors. \emph{Adequacy} \cite{d2019automatic,ma2017semantic,mcburney2014automatic,arumae2019guiding,libovicky2018multimodal} rates the amount of meaning expressed in the generated sample given a reference sample. Human participants and categorical scales dominate the assessment process. \emph{Topicality} expresses how well does the generated sample topic match one of the reference samples \cite{riedhammer2008keyphrase,arumae2019guiding,fang2017word}. \emph{Naturalness} shows the likelihood of a text being natural or written by a human being rather than automatically generated \cite{ccano2020human}.
\emph{Relevance} represents how closely are the documents related \cite{bhatia2014summarizing,Erol03multimodalsummarization,murray2010generating,DBLP:journals/corr/abs-2004-02016,zhang2012automatic,zhu2020hierarchical,lee2020reference}. \emph{Consistency} represents the degree of agreement with the original content \cite{kryscinski2019neural,wang2020asking,lee2020reference}. \emph{Fluency} represents the quality of expression \cite{oya2014,wang2013domain,oya2014template,lee2020reference}. \emph{Coverage} determines how much of the important content is covered from the source document in the summary \cite{sonjiaLevow,gillick2009global,li2019keep,mehdad2013abstractive}. \emph{Informativeness} represents the importance of the content captured in the summary \cite{zhang2020unsupervised,liu2009extractive,oya2014template,oya2014}. Besides accuracy, the rest of the above quality criteria are assessed manually by human experts or survey participants \cite{zhu2006summarization,shirafuji2020summarizing}.

\subsection{Automatic Evaluation}
\label{subsec:automatic_evaluation}
The current automatic evaluation of various text summarization tasks (including minuting) is mostly based on ROUGE or similar metrics that utilize n-gram comparisons (from single words to long patterns). While automatic and fast, these metrics are often not able to reflect the quality issues of the text samples \cite{see-etal-2017-get}. Some of the typical problems they miss are grammatical discrepancies, word repetitions, and more. \protect\newcite{novikova-etal-2017-need,reiter-2018-structured} also report that automatic metrics do not correlate well with human evaluations. To overcome these limitations, it is important to simultaneously run human evaluations (following a systematic protocol) of meeting summaries and augment the automatic metric scores with the manual ones. 

\section{The \toolname{} Annotation Tool}

\begin{figure*}[t]
    \centering
    \includegraphics[width=0.97\textwidth]{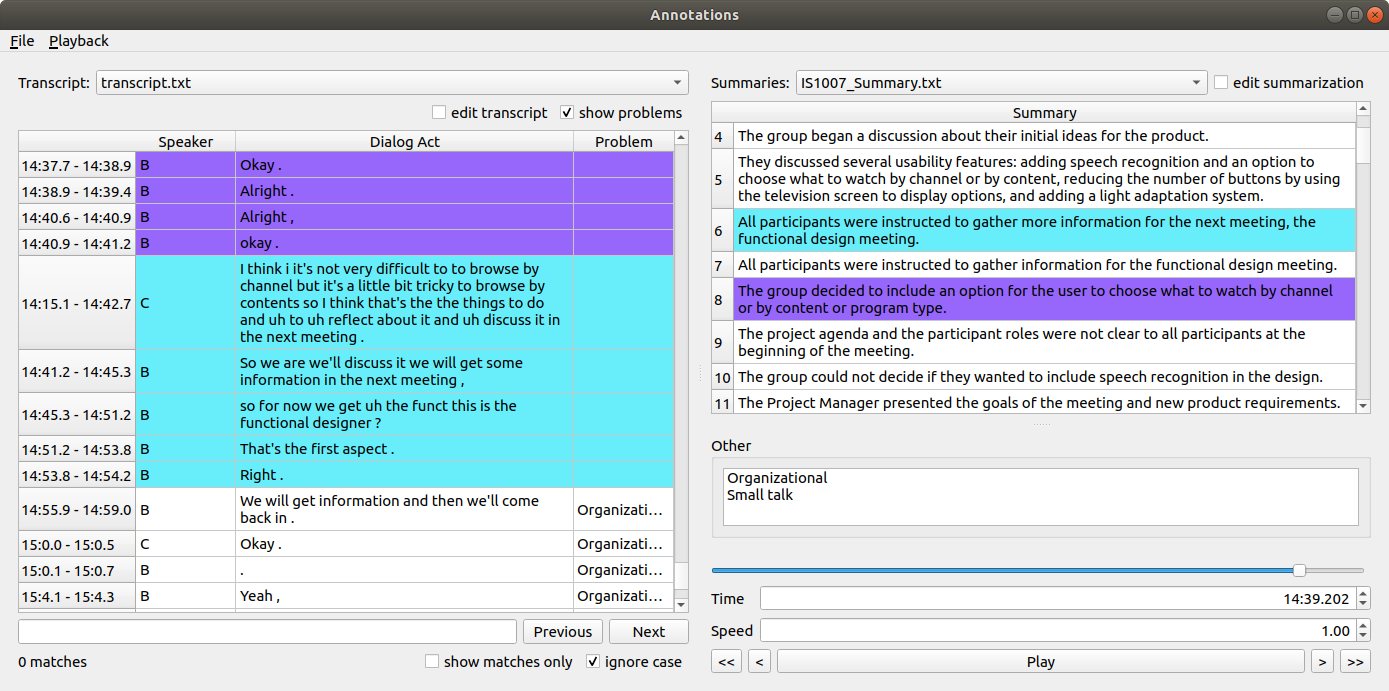}
    \caption{The \toolname{} main view  in the annotation mode. The left column contains the meeting transcript broken down to dialogue acts. The right column contains a summary, and the player. The alignment between dialogue acts and the summary point is shown using colors.}
    \label{fig:annotation_mode}
\end{figure*}

\begin{figure*}[t]
    \centering
    \includegraphics[width=0.97\textwidth]{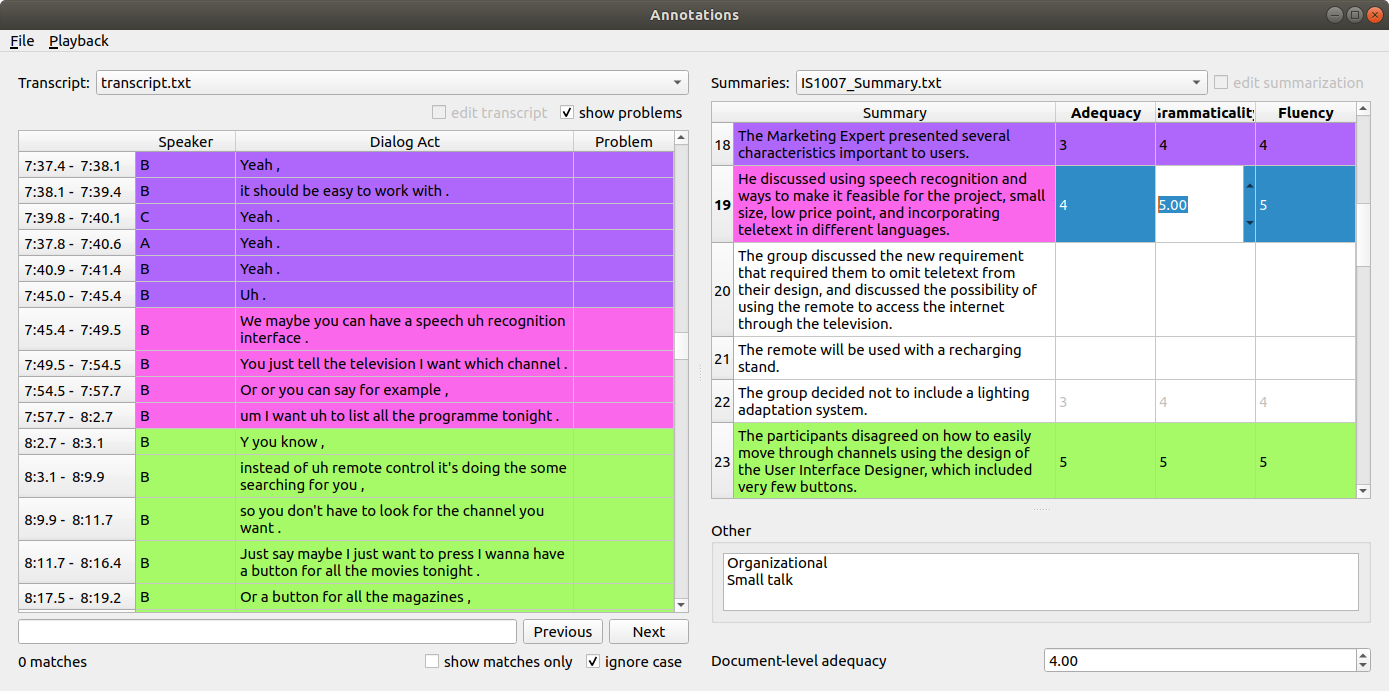}
    \caption{The \toolname{} main view in evaluation mode. The left column contains the meeting transcript broken down into dialogue acts. The right column contains a summary, problem flags, and document-level adequacy. Evaluation, i.e., the assignment of scores to a particular summary point, is enabled only for the summary points where the corresponding DAs are visible in the transcript view.}
    \label{fig:evaluation_mode}
\end{figure*}

\toolname{} is a comprehensive annotation and evaluation tool. It supports all stages of the preparation and/or evaluation of a corpus of multi-party meetings, i.e., creation and editing of meeting transcripts, annotating speakers, creating a summary, alignment of meeting segments to a summary, and meeting summary evaluation.

The tool is written in Python using PySide\footnote{\url{https://www.qt.io/qt-for-python}} for GUI which makes the tool available on all major platforms (i.e., Windows, Linux, and macOS). 

\subsection{Design Choices}
We represent a meeting with its transcript and summary in \cref{fig:annotation_mode}. The transcripts are long documents consisting of multi-party dialogues (refer to the left side of the tool window). The meeting summary is a structured document. We decided to break down the meeting summary into separate \emph{summary points}. A summary point roughly represents a line in a summary document (refer to the right part of the tool window).
The meeting usually has more versions of transcripts (e.g., generated by ASR and a manual one) and more versions of summaries (e.g., supplied by meeting participants created during the meeting and others provided by an annotator). We add drop-down lists to select a specific version of the transcript and summary. If the user changes the version of one, the program loads the appropriate version of the alignment automatically.

We segment the transcript into dialogue acts (DAs). A DA represents the meaning of an utterance at the level of illocutionary force \cite{austin1975things}. In the context of our tool, a DA represents a continuous portion of a transcript uttered by one speaker on a single topic. We believe that for better readability, the DA might be further broken down into smaller units. 

Optionally, the meeting might have an audio or video recording. The meeting recording is helpful during the meeting annotation (i.e., creating/editing the meeting transcript and summary). The tool offers an embedded player. Then, the annotator does not have to switch between the annotation tool and a media player. Also, if the transcripts come with timestamps, the annotator can easily seek in the player by double-clicking the particular DA.

Many annotation tools we reviewed in \cref{tools} provide automated suggestions. We decided not to include this feature as we believe it would bias the annotators.

\toolname{} is designed with two modes: Annotation and Evaluation. We further elaborate them in \cref{annotation_mode,evaluation_mode}.

\subsection{Annotation}
\label{annotation_mode}
The annotation task consists of several sub-tasks. We envision the following sub-tasks: (1) transcript annotation, (2) summary annotation, and (3) alignment. 

\subsubsection{Transcript Annotation}
Transcripts may be either generated by an ASR or manually created. The tool supports both scenarios, i.e., transcribing the recording, post-editing, splitting the transcript into dialogue acts, and speaker annotation.

We introduce a set of keyboard shortcuts that make simple tasks like creating/deleting or even splitting DAs very quick. Additionally, we offer a search toolbar supporting regular expressions.%

\subsubsection{Summarization Annotation}
Summarization annotation involves the creation or possible modification of an existing meeting summary. The tool provides a convenient platform to add more points to an existing summary by simply clicking the ``add" or ``delete" buttons.

Except for summary points, we intentionally do not enforce any precise summary structure and provide users with the flexibility to design their summary. 
Though, we support indentation as a form of horizontal structuring (with a user-defined indentation symbol).

\subsubsection{Alignment}
The alignment captures which dialogue acts are associated with a particular summary point. We call a set of DAs belonging to a summary point a \textit{hunk}. 
DAs which do not correspond to a summary point may be assigned meta-information (i.e., marked as small talk or organizational).

\toolname{} supports only $n$-to-1 alignments because we believe that aligning multiple summary points to a DA would further increase the difficulty of the alignment task. It would also cause a ``summary point fragmentation'', as the annotator might address the same information in separate summary points. When a DA includes more information that fits in a single summary point, we suggest splitting the DA accordingly.

The matching background color of a hunk and a summary point represents a single alignment (see \cref{fig:annotation_mode}). To make the interface more clean and readable for the annotator, we color only summary points whose hunks are currently visible in the transcript view.

Aligning DA(s) to a particular summary point or meta-information item is very intuitive: 

\begin{enumerate}[itemsep=-1mm]
    \item \textit{Select DA(s)} in the transcript view. The selection can be contiguous and also discontiguous. Standard GUI gestures are supported (i.e., dragging over items, \texttt{[Ctrl]/[Shift]} + clicking/dragging).
    \item \textit{Select a summary point} by double-clicking an item in the summary view or choose a \textit{problem label} in the meta-information list.
\end{enumerate} 

Resetting the alignment is also possible by selecting DA(s) or a summary point and selecting an action in the context menu or keyboard shortcut.

In this way, \toolname{} facilitates the annotation and mitigates potential errors. The annotator has a clear overview of which parts of a meeting are already annotated and makes any revisions straightforward.

\subsection{Evaluation Mode}
\label{evaluation_mode}
We reviewed several quality criteria for a summary evaluation in \cref{subsec:manual_evaluation,subsec:automatic_evaluation} based on which we formulate a novel manual evaluation scheme. We integrated the evaluation into the tool (see \cref{fig:evaluation_mode}).

For the evaluation, we utilize \emph{adequacy}, \emph{grammaticallity} and \emph{fluency}. We think that evaluating these criteria at the document level is challenging and error-prone. Therefore, we propose the evaluation on two levels: (1) manually assigning the hunk level (based on alignment) and (2) automatically aggregating it on the document level. At the hunk level, the evaluation is based only on the aligned part of the transcript and a corresponding summary point. 

At the hunk-level, annotators evaluate \emph{adequacy}, \emph{grammaticality} and \emph{fluency}  using a 5-star scale \cite{likert} with 1 being the worst and 5 the best.
At the document level, we %
automatically aggregate the hunk-level judgments with a simple average.
Aside from averaging hunk-level adequacy across the document, we also independently ask annotators to report the overall accuracy of the minutes. We call this score `\emph{Doc-level adequacy}' in the following.
Finally, we compute \emph{coverage}, i.e., the number of aligned DAs divided by the total number of DAs.

\section{Use Case and Pilot Study}

In this section, we present a use case and conduct a small-scale pilot study.

\subsection{Use Cases}
We organized the First Shared Task on Automatic Minuting \cite{automin} on creating minutes from multi-party meetings. As a part of the shared task, we made available a minuting corpus, which is now being released publicly \cite{elitr-minuting-corpus:2022}.
\toolname{} was created during the annotation process. We have started with a modified NITE \cite{kilgour2006nite} tool, but the annotators faced many issues, including the need to make changes to the transcript and minutes. Hence, we decided to create a new tool to meet the annotators' requirements. We used agile development, i.e., we constantly improved \toolname{} following the annotators' comments.

Before annotation, each meeting consisted of a recording, ASR-generated transcript, and meeting minutes assembled by the meeting participants (often incomplete). First, we asked the annotators to revise the ASR transcript. Later, we asked the annotators to provide minutes and alignment. We have observed different styles of minuting among the annotators. Therefore, many of the meetings have two or more versions of minutes provided by different annotators.

\subsection{Pilot Study}
To assess \toolname{}, we conduct a simple experiment similar to \protect\newcite{collins2019lida} for both modes of tool: (1) annotation and (2) evaluation. 
We evaluate all the results across two different meeting corpora, AMI \protect\cite{mccowan2005ami} for English and AutoMin for Czech.  We considered one meeting per language from each corpus (the selected English meeting has 205 DAs %
and the selected Czech meeting has 153 DAs; both are approximately 16 minutes long). The task was to create an abstractive summary, align the transcript with the corresponding parts of the reference summary, and finally evaluate the reference summary relying on the constructed alignment. Each of the three annotators had a different experience level and report their timings in \cref{tab:results}.  %
The summarization stage took on average 40.7 minutes and 33.0 minutes for English and Czech, respectively. The alignment took on average 16.0 and 19.7 minutes and evaluation on average of 11.7 and 17.7 minutes for English and Czech data, respectively. In other words, this particular meeting needed about 2--3 times its original time to summarize, its duration to align, and finally somewhat less than its duration to evaluate. Based on this minimal study, a factor of 4 or more has to be expected when processing meetings by annotators who have not taken part in them. 

The evaluation results are in \cref{tab:results2}. Adequacy is deemed average (3.98$\pm$0.62 on average), with the document-level manual judgment being similar (3.83$\pm$0.37), while grammaticality and fluency are somewhat higher (4.32$\pm$0.39 and 4.63$\pm$0.31, resp.). Additionally, we report the inter-annotator agreement (IAA). Our definition of IAA is rather strict, we count the number of DAs 
that were aligned to the same summary point by all annotators divided by the total number of DAs. 

If we consider the recorded pace of our annotators, the AMI meeting corpus consisting of 137 meetings and 45,895 DAs in total (see \cref{tab:comparative}), it would take 9,105 minutes to summarize, 3,582 minutes to align, and 2,613 minutes to evaluate using our tool, or 255 hours in total. We infer from \cref{tab:results} that the time spent on the task does not necessarily depend on the annotator's experience but rather on the personal preferences and thoroughness of the annotator. Despite the limited size of the experiment, we believe that the results suggest the tool is intuitive and facilitates fast annotation. 

\begin{table}[t]
\centering
\resizebox{0.4\textwidth}{!}{%
\begin{tabular}{l||ccc||ccc}
            & \multicolumn{3}{c||}{English} & \multicolumn{3}{c}{Czech}  \\ \hline
Annotator     & E1   &  E2 & E3  & C1   & C2 & C3  \\ \hline
Experienced     & \protect\xmark   &  \ding{51} & \ding{51}  & \protect\xmark   & \ding{51} & \ding{51}  \\ \hline
Summarization   & 37       &   45     &  40       & 23       & 45     &  31     \\
Alignment       & 5        &   23     &  20        & 18       & 30     &  11      \\
Evaluation      & 10       &   15     & 10        & 25       & 15     &  13      \\ \hline
Total time      & 52       &   83      & 70        & 66       & 90     &  55      
\end{tabular}%
}
\caption{Pilot study: annotator experience and time in minutes each annotator spent on each task.}
\label{tab:results}
\end{table}

\section{Conclusion}
We presented \toolname{}, an open-source and intuitive comprehensive tool for meeting annotation. Its main goal is to facilitate alignment between parts of a transcript with the corresponding part of the summary. We also integrate the proposed evaluation strategy of meeting summaries in the tool. 

In the future, we will add the support for automatic transcript generation with timestamps, user-defined problems in the list of explicit problem labels, and a quick onboarding tutorial integrated into the user interface. Finally, we hope \toolname{} will generally improve as annotators will provide their feedback. 

\begin{table}[t]
\centering
\resizebox{0.49\textwidth}{!}{%
\begin{tabular}{l||ccc||ccc}
            & \multicolumn{3}{c||}{English} & \multicolumn{3}{c}{Czech}  \\ \hline
Annotator     & E1   &  E2 & E3  & C1   & C2 & C3  \\ \hline
Experienced     & \xmark   & \ding{51} & \ding{51}  & \xmark& \ding{51} & \ding{51}       \\ \hline
\#Summary points & 15       &  11       & 19      & 23       & 14     & 21       \\ 
\#Alignments    & 378      &   378     & 203      & 282      & 282    & 282       \\ \hline
IAA    & \multicolumn{3}{c||}{0.21*} & \multicolumn{3}{c}{0.63}  \\ \hline
Avg. adequacy          & 3.71      & 3.71      & 3.17     & 3.67     & 4.93    & 4.67   \\ 
Avg. grammaticality    & 3.86      & 4.21       & 4.08   & 5.00      & 4.13    & 4.67   \\ 
Avg. fluency           & 4.71      & 4.07       & 4.92     & 5.00      & 4.53    & 4.53   \\ 
Doc.-level adequacy    & 3.00      & 4.00       & 4.00    & 4.00      & 4.00    & 4.00    \\ \hline
Coverage   & 1.00      &   0.94     & 0.54      & 0.64      & 0.54    & 0.30 \\
\end{tabular}
}
\caption{Pilot study: annotator experience, number of produced summary points and alignments, and evaluation score.\\
$*$ If we remove the second annotator, we obtain agreement 0.59.}
\label{tab:results2}
\end{table}

\section*{Acknowledgements}
This work has received support from the project ``Grant Schemes at CU'' (reg. no. CZ.02.2.69/0.0/0.0/19\_073/0016935), the European Union's Horizon 2020 Research and
Innovation Programme under Grant Agreement No 825460 (ELITR), and  19-26934X (NEUREM3) of the Czech Science Foundation, and partially supported by SVV project number 260 575.

\section{Bibliographical References}\label{reference}

\bibliographystyle{lrec2022-bib}
\bibliography{custom}

\end{document}